Tkaczyk, Dominika, Andrew Collins, and Joeran Beel. "NaïveRole: Author-Contribution Extraction from Biomedical Publications." In *27th AIAI Irish Conference on Artificial Intelligence and Cognitive Science*, 2019.

# NaïveRole: Author-Contribution Extraction and Parsing from Biomedical Manuscripts

Dominika Tkaczyk [1,a], Andrew Collins[1,b], Joeran Beel[1,c]

[1] Trinity College Dublin, ADAPT Centre, School of Computer Science and Statistics, Ireland
[a]d.tkaczyk@gmail.com, [b]ancollin@tcd.ie, [c]joeran.beel@scss.tcd.ie

**Abstract.** Information about the contributions of individual authors to scientific publications is important for assessing authors' achievements. Some biomedical publications have a short section that describes authors' roles and contributions. It is usually written in natural language and hence author contributions cannot be trivially extracted in machine readable format. In this paper, we present 1) A statistical analysis of roles in author contributions sections, and 2) *NaïveRole*, a novel approach to extract structured authors' roles from author contribution sections. For the first part, we used co-clustering techniques, as well as Open Information Extraction, to semi-automatically discover the popular roles within a corpus of 2,000 contributions sections from PubMed Central. The discovered roles were used to automatically build a training set for *NaïveRole*, our role extractor approach, based on Naïve Bayes. *NaïveRole* extracts roles with a micro-averaged precision of 0.68, recall of 0.48 and F1 of 0.57. It is, to the best of our knowledge, the first attempt to automatically extract author roles from research papers. This paper is an extended version of a previous poster published at JCDL 2018.

## 1 Introduction

Authorship is an important concept in scholarly communication. It allows people to properly credit those who contributed to scientific discoveries and is widely used to assess people's scientific achievements. However, to fully evaluate researcher's achievements, it is useful to know the precise nature of their contributions to authored publications. In some biomedical journals, a submitting author must provide information about each author's individual contributions. This information is then attached to the manuscript as a short section entitled e.g. "Authors' Contributions" (**Fig. 1**). Examples of contributor roles include the preparation of data, designing experiments, programming software, or writing and editing the manuscript.

These sections are usually written in natural language, are unstructured, and are intended for humans to read rather than machines. Contribution taxonomies and machine-readable formats are being introduced slowly, however, digital libraries contain documents that have already been published in previous decades. Contribution information in such documents will not conform to new standards and will remain in an unstructured format. Consequently, analyses of author contribution information requires time-consuming manual work, which makes processing large collections of documents in digital libraries impractical. We address these issues by proposing:

1. a method for semi-automatically discovering what roles are common in a corpus of sections of interest
2. a scalable approach for annotating a ground truth role dataset
3. a supervised algorithm for automatic extraction of the roles from unstructured text

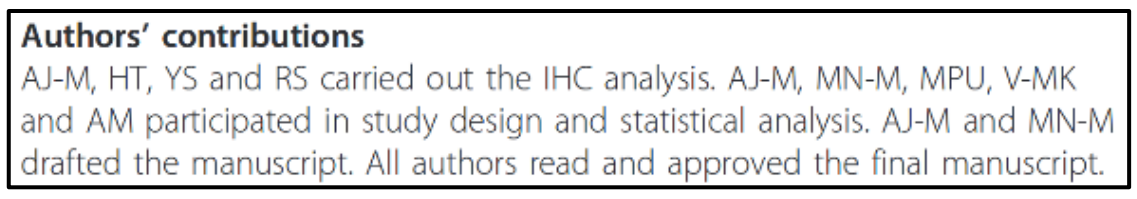
**Authors' contributions**
AJ-M, HT, YS and RS carried out the IHC analysis. AJ-M, MN-M, MPU, V-MK and AM participated in study design and statistical analysis. AJ-M and MN-M drafted the manuscript. All authors read and approved the final manuscript.

**Fig. 1.** Example of "Authors' contributions" section with abbreviated author names.

This paper is an extended version of a poster published at the Joint Conference on Digital Libraries [1]. This extended version contains further descriptions of our study and proposed approaches, a comparison of our results to an existing contributor role taxonomy, and an error analysis of the proposed automatic role extractor.

## 2 Related Work

General information extraction from scientific literature is a popular research area, resulting over the years in many approaches and tools, including CERMINE [2], GROBID [3], PDFX [4], ParsCit [5], Science Parse[1] and Docear PDF Inspector [6]. However, none of these systems extracts information related to the contributions of individual authors directly from the content of the paper.

The scientific community has thus far not agreed on standard author contributions or even standard criteria for authorship. Nevertheless, some initiatives have been undertaken to increase the level of consistency between journals. For example, the International Committee of Medical Journal Editors published guidelines that suggest minimum requirements for authorship, and the use of these guidelines is now encouraged by some medical journals.

CRediT [2] is an example of a contribution taxonomy that defines the standard for contributors' roles. CRediT is composed of 14 roles and was created based on free-form contributions and acknowledgements sections. Journals are increasingly adopting taxonomies like CRediT to consistently describe author contributions [8]. Our study does not assume any input taxonomy but aims at discovering popular roles within a corpus of contribution descriptions in an unsupervised way.

Some journals, such as PLOS One or Annals of Internal Medicine, publish author contribution information in a machine-readable form. Several studies have examined author contributions using this data, for example, comparing author orderings to contributions [9, 10]. Typically, however, author contribution information has an unstructured, natural language form, and cannot be trivially examined in this fashion.

[1] https://github.com/allenai/science-parse (we used version 1 as at the time of our analysis the currently released version 2 was not available, or we were at least not aware of it)

[2] Contributor Roles Taxonomy: https://casrai.org/credit/

# 3 Methodology

## 3.1 Roles Discovery

The first stage of our workflow (**Fig. 2**) is to discover common roles appearing in the corpus. Our analysis was composed of the following steps:

- Data preparation, where we gathered a corpus of contributions sections.
- Data preprocessing, where role mentions were extracted and cleaned.
- Clustering, where abstract role concepts were discovered.

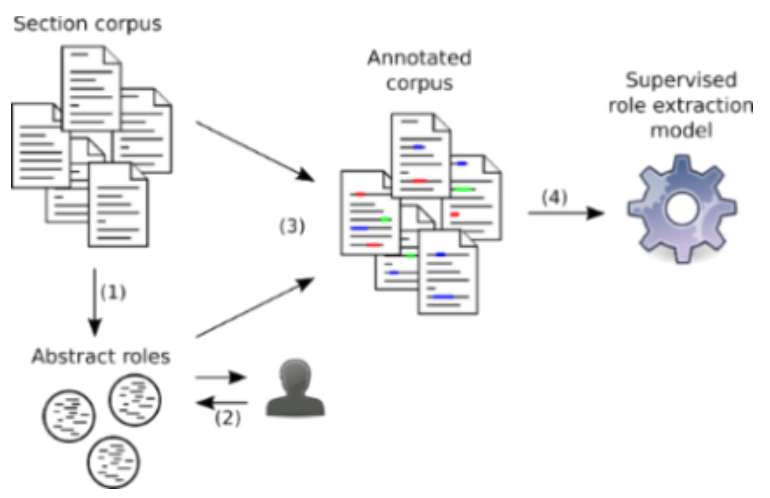

**Fig. 2.** The workflow of our study. First, role mentions from the corpus were clustered to discover abstract role concepts (1). Then, resulting clusters were manually inspected and corrected (2). Next, cleaned clusters were used to generate the training set (3). Finally, a supervised machine learning model able to classify role mentions was trained (4).

**Data Preparation**

We use the PubMed Central Open Access Subset as data for our work. This is a subset of the total collection of articles in PMC, published under open licenses. We downloaded the corpus of 1.6 million documents in machine-readable JATS format[3]. From each document we extracted any section whose normalized (lowercased and with all non-letters removed) title equals “authorscontributions”. We found these sections in 186,874 documents, constituting ~12% of the corpus. For performance reasons, we use a random subset of 2,000 sections only. All sections are written in English.

**Preprocessing**

Authors’ contribution sections typically mention the roles of several individual authors. We refer here to a natural language expression of the role an author plays as a *role mention*. A same role (e.g. *data analysis*) can be expressed by many forms of role mention (e.g. “X analyzed microarray sequences”, “X was involved in data analysis”).

[3] https://jats.nlm.nih.gov/

We represent a role mention as a 3-element tuple containing: 1) subject: “who”, usually author name or initials, 2) action: activity, often a verb phrase, 3) object: “what the action was applied to”, typically a noun phrase (**Fig. 3**).

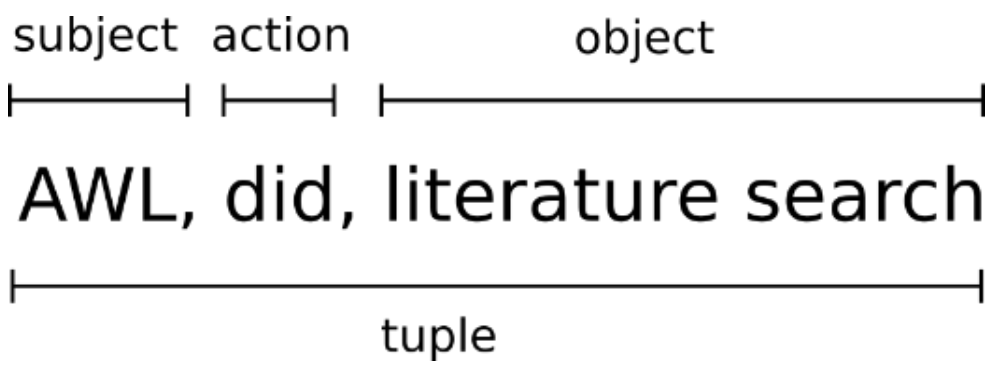

**Fig. 3.** The decomposition of a single role mention into three parts: subject, action, and object.

We use the Stanford Open Information Extraction tool[4] to extract role mentions from the text. OpenIE [11] is an information extraction paradigm, in which it is possible to extract relations in the form of tuples (relation plus its two arguments) from the text, in an unsupervised way. The output corresponds to 3-element role mentions, where *action* is the relation expression and *subject* and *object* are its two arguments.

As a result of applying OpenIE to our sections corpus, for every section we obtained a bag of role mentions, where a mention is a tuple of three text fragments from the original text. For example, from the sentence “AWL did the literature search and participated in the writing of the manuscript.” we got the following tuples: (“AWL”, “did”, “literature search”) and (“AWL”, “participated in”, “writing of manuscript”).

OpenIE tools tend to output tuples that are redundant. For example, from the same sentence we might get both (“authors”, “read”, “final manuscript”) and (“authors”, “read”, “manuscript”) tuples. We analyze all pairs of tuples and consider one tuple in a pair redundant if the following conditions were met: 1) their subjects are exactly the same, 2) the action of one tuple contains all the words of the other action in the same order, and 3) the object of one tuple contains all the words of the other object in the same order. We remove such redundant tuples.

The roles in role mentions are expressed by action-object pairs, and the subject refers only to the author. At the beginning, our corpus of 2,000 sections contained 6,924 distinct action-object pairs, many of which expressed the same roles.

To merge some mentions and reduce the number of distinct action-object pairs, we applied cleaning and normalizing to actions and objects of role mentions. First, we stemmed words within actions and objects, and removed stopwords. For stemming we used R’s SnowballC library, and the stopwords list was downloaded from an online source[5]. This reduced the number of distinct roles to 6,289. We also remove rare role mentions, that is, mentions appearing less than five times in the corpus. This leaves 434 distinct action-object pairs while keeping 55% of role mentions.

Finally, we observed that due to splitting role mentions into action and object, we still have distinct mentions that obviously refer to the same role, such as (“analys”, “data”) and (“perform”, “the analys of the data”). We wanted to normalize this, at the same time keeping the tuple-based structure of the mentions. To achieve this, we extracted a number of most common terms from both actions and objects of the mentions (terms appearing at least 20 times in the corpus), and then each term was labeled as "action keyword" or "object keyword", based on whether it is more common among actions or objects.

[4] https://nlp.stanford.edu/software/openie.html

[5] http://www.ranks.nl/stopwords

**Table 1** lists extracted action and object keywords. Each role mention in the corpus was then transformed in the following way: 1) the subject was left intact, 2) all action keywords found in the entire original mention formed the new action, and 3) all object keywords found in the entire original mention formed the new object. In addition, if the new action turned out to be empty, we added a single "perform" keyword to it.

**Table 1.** Action and object keywords appearing in the corpus. The words are stemmed.

| Action keywords | Object keywords |
|---|---|
| read, particip, draft, contribut, conceiv, perform, write, revis, carri, critic, approv, made, prepar, conduct, provid, review, supervis, equal, develop, edit, plan, initi, acquir, assist, coordin, help, took, undertook, gave, comment, take, recruit | manuscript, studi, data, final, design, analys, experi, collect, interpret, statist, respons, involv, paper, concept, result, version, substanti, acquisit, project, patient, research, work, content, intellectu, import, articl, discuss, first, protocol, molecular, investig, sequenc, literatur, idea, part, princip, clinic, trial, sampl, genet, laboratori, advic, tool |

This operation moved words between actions and objects so that action keywords are always in the actions of the mentions and object keywords are in their objects. For example, since "perform" is an action keyword, and "analys" and "data" are object keywords, both mentions ("analys", "data") and ("perform", "the analys of the data") became ("perform", "data analys"). This process left us with 285 distinct role mentions.

### Finding Roles

In this phase, we detect roles in our collection of role mentions. We adopted an unsupervised machine learning technique (clustering) for this task. This is similar to a standard ontology learning approach [12]. At the end of clustering, all mentions that refer to the same role should belong to the same cluster. For example, ("performed", "data analysis") and ("was involved in", "analyzing data") should be clustered together. After preprocessing, our set contained 9,709 role mentions represented by cleaned subject-action-object tuples. We were interested in co-clustering the actions and the objects separately yet simultaneously, which in turn would define a third clustering based on the combinations of actions and objects.

More formally, let $M = \{m_1, \dots, m_N\}$ be the input mention set, and $A$ and $O$ the set of action clusters and the set of object clusters, respectively. We define an action clustering as a function $f_a: M \rightarrow A$, which maps mentions to their action clusters. Similarly, let $f_o: M \rightarrow O$ be the mapping function which defines object-based clustering. This lets us define a role set $R$ as the set containing all combinations of action and object concepts that share some mentions: $R = \{(a, o) \in A \times O \mid f_a^{-1}(a) \cap$ $f_o^{-1}(o) \neq \emptyset\}$. The final combined clustering is $f_r: M \to R$ such that $\forall_{m \in M} f_r(m) = \big(f_a(m), f_o(m)\big)$.

Set $R$ defines a binary relation between action and object clusters. We can define the weight of this relation as the number of the mentions that the clusters share: $\forall_{a \in A, o \in O} r(a,o) = |\{m \in M \mid f_r(m) = (a,o)\}| = |f_r^{-1}(a,o)|$. Intuitively, if an action concept and an object concept appear in many role mentions together, they form a common role, and the weight of the role is large. This defines a graph structure among the clusters, with action and object concepts as nodes and weighted edges representing relation strength.

Finally, during our analysis we used the idea of a cluster label, defined as a bag of terms of the most numerous member of the cluster.

We use bottom-up clustering, where we start with initial action and object clusters, and in several phases we merge clusters together. Initially, the clusters are defined as distinct normalized actions and objects. In other words, two mentions are in the same action/object cluster if their normalized actions/objects are identical. Each round of clustering is composed of two stages. The first one is based purely on cluster term labels. The second one uses the graph structure defined previously. Algorithm 1 presents the pseudocode of the role mentions clustering.

The first stage of the clustering is based on the action/object label terms of the current role clusters. We examine pairs of role clusters and merge them if action and object terms of one of them contain the other cluster's terms. The new cluster is always given a label equal to the label of the bigger cluster from the examined pair.

The main clustering stage is based on the weighted graph relations between action and object clusters. First, we identify an action or object cluster pair that is most similar to each other, then their clusters are merged. When the highest similarity drops below a predefined threshold, the clustering procedure terminates. We will only explain how the similarity between two action clusters is defined. The similarity between object clusters is defined analogously.

The main observation used for calculating the similarity between two action clusters is that two actions related to a lot of common objects will be more similar to each other. However, this assumption is trivially violated in cases where there simply are different ways we can affect the same object (for example the manuscript can be read, written, reviewed, etc.). In such cases we would like the overall similarity to be lower.

**Algorithm 1:** Role mentions clustering

```
action_clusters ← grouping of actions by their normalized value
object_clusters ← grouping of objects by their normalized value
similarity ← ∞
while similarity > threshold do
   for each role cluster pair do
      if one element contains all terms of the other then
         merge clusters & relabel the smaller cluster
      end
   end
   pair ← action or object cluster pair with the highest similarity
   similarity ← the highest similarity
   merge clusters from pair & relabel the smaller cluster
```

To reflect these observations, we introduce an object weight which is the reciprocal of the number of distinct actions it is related to: $\forall_{o \in O} w(o) = |\{a \in A \mid (a, o) \in R\}|^{-1}$. Intuitively, an object with a small weight (such as “manuscript”) interacts with many different actions, in other words there are many actions that can be applied to it.

We define the similarity between two actions as the sum of the weights of all the objects they share: $\forall_{a_1, a_2 \in A} s(a_1, a_2) = \sum_{o \in O, (a_1, o) \in R, (a_2, o) \in R} w(o)$. Intuitively, two actions will have high similarity if: 1) they share a lot of objects, and 2) the objects they share are “specific” (few distinct actions apply to them). An object that interacts with many actions will not contribute much to the action similarity.

Examples of merged clusters include: “particip” and “perform”, “contribut” and “perform”, “assist” and “perform”, “manuscript” and “paper”, “carri” and “perform”, “experi” and “study”, “perform” and “undertook”, “manuscript” and “articl”.

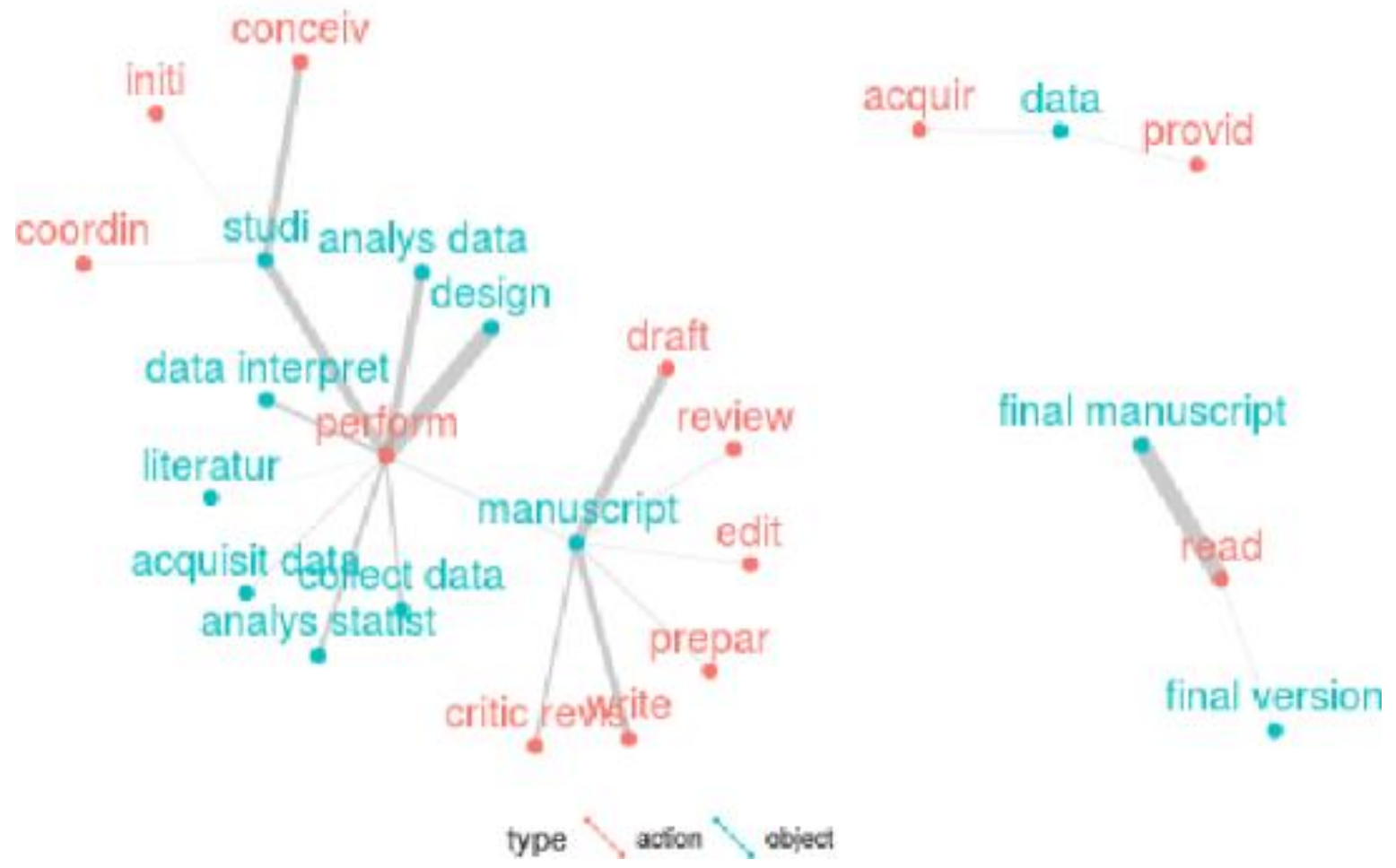

**Fig. 4.** The role graph resulting from automated clustering. The nodes represent action and object clusters (their labels are bags of stemmed terms). The width of edges represents the strength of the relation between action and object nodes. Less common roles were removed.

The clustering procedure resulted in reducing the number of role clusters from 285 to 63. The following clusters were merged:

1. “particip” and “perform”
2. “contribut” and “perform”
3. “assist” and “perform”
4. “manuscript” and “paper”
5. “project” and “study”
6. “carri” and “perform”
7. “experi” and “study”
8. “perform” and “undertook”
9. “manuscript” and “articl”
10. “approv” and “read”
11. “made” and “perform”
12. “conduct” and “perform”
13. “perform” and “supervis”
14. “help” and “perform”
15. “perform” and “plan”

The procedure made a few errors, merging for example: "approv" and "read", "perform" and "supervis". The final graph is shown in **Fig. 4**.

### 3.2 Manual Correction

To reduce the number of errors from automatic clustering, we manually inspected 63 clusters. This included removing some clusters and merging others. We also assigned role names to the clusters. The entire procedure resulted in 13 roles. The final set of 13 roles, as well as the fractions of mentions for every role, are presented in **Fig. 5**.

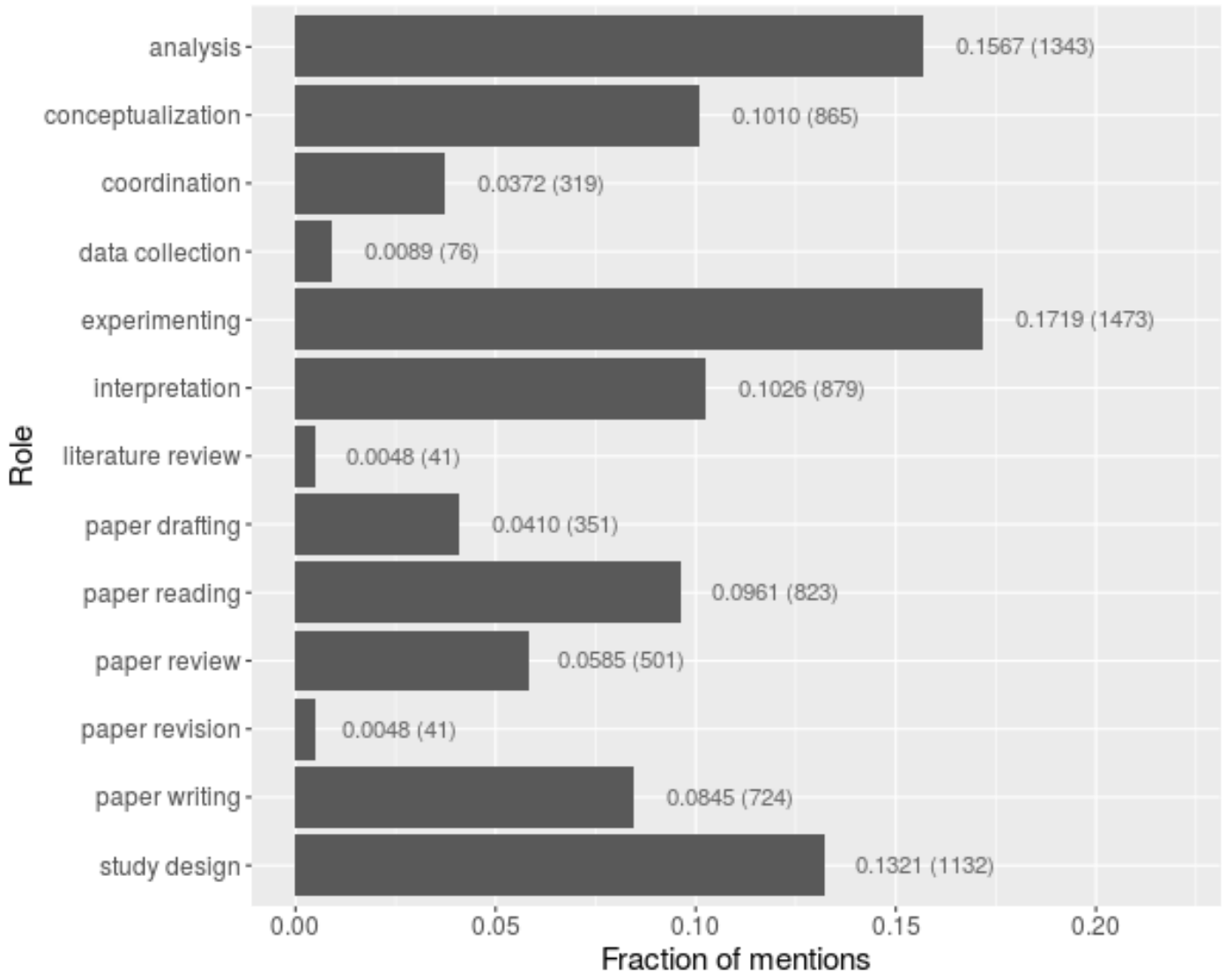

**Fig. 5**: The final set of roles, showing the counts and fractions of the entire role mention set.

### 3.3 Annotating the Dataset

We annotated the dataset of role mentions. More specifically, the dataset contains role mentions labelled with abstract roles. For example, the dataset might contain the entry: ("participated in, the analysis of microarray data", data analysis). The resulting dataset is composed of the role mentions from the clusters, and the label for each role mention is the role name assigned to the mention's cluster. This annotation approach differs from the typical approach, in which we would manually label each role mention in the dataset. Even though our approach still requires manual work, it was performed on the clusters, not each individual role mention. Since the clusters are much less numerous than the role mentions, our proposed approach is less labor intensive.

### 3.4 Roles Extraction from the Text

This section describes our prototype of an automated extractor of authors' roles from text. The extractor takes a contributions section as input and outputs a set of extracted roles. We used the previously developed preprocessing pipeline and discovered roles for this task. The extraction algorithm is composed of the following steps:

- First, a set of role mentions is extracted from the text of the section. If the section is written in a natural language, this is done using OpenIE. In some rare cases we came across, the contributions section was not written in natural language, but rather contained a list of contributions in the following format (or a variation of it): "author1: role1, role2; author2: role3; ...". In such cases we extract role mentions using regular expressions. Redundant mentions are then removed.
- Next, each mention is represented as a feature vector. We use a binary bag-of-words representation, with 64 words corresponding to the object/action keywords (**Table 1**). Only the keywords that remained after manual cluster removal are used.
- Finally, each mention is classified by a supervised Naïve Bayes model trained on the mention set generated previously. The final output is a set of author-role pairs.

## 4 Results

### 4.1 Roles Discovery

**Fig. 4** shows the graph resulting from the automated clustering procedure (before manual correction). The final corrected roles resulting from our study are: *experimenting* (1,743 instances, 17% of the entire role set), *analysis* (1,343, 16%), *study design* (1,132, 13%), *interpretation* (879, 10%), *conceptualization* (865, 10%), *paper reading* (823, 10%), *paper writing* (724, 8%), *paper review* (501, 6%), *paper drafting* (351, 4%), *coordination* (319, 4%), *data collection* (76, 1%), *paper revision* (41, 0.5%) and *literature review* (41, 0.5%).

Our final role set was manually compared to the existing taxonomy CRediT. It is important to note that our study was based on biomedical data only, while CRediT is a general-purpose taxonomy. As a result, some differences are to be expected.

**Table 2.** Comparison of the roles discovered by our study and existing taxonomy CRediT.

| Similarities | | Differences | |
|---|---|---|---|
| Our study | CRediT | Our study | CRediT |
| Analysis | Formal analysis | Paper reading | - |
| Conceptualization | Conceptualization | Literature review | - |
| Experimenting | Investigation | Interpretation | - |
| Study design | Methodology | - | Software |
| Coordination | Project administration | - | Validation |
| Data collection | Resources | - | Funding acquisition |
| Paper drafting/ Paper writing | Writing – original draft | - | Supervision |

| | |
|---|---|
| Paper review/<br>Paper revision | Writing - review & editing |

In general, the results are similar (**Table 2**). Five roles appear in both our clusters and CRediT. Our study resulted in four roles related to preparing the manuscript itself, while CRediT has only two such roles. Three roles discovered in our study (*paper reading*, *literature review* and *interpretation*) are not included in CRediT.

### 4.2 Roles Extraction

To evaluate our role extractor, we manually annotated a test set of 100 contributions sections. At this point, we observed three new roles that were not discovered in our study: *paper approving*, *supervision* and *funding acquisition*. Since the classifier does not have any training data for these roles, they are never assigned.

During the evaluation, for every document we compared the extracted author-role pairs to the ground truth pairs. A pair was marked as correctly extracted if identical to any pair in the ground truth. We obtained the following micro-averaged results: precision 0.68, recall 0.48, F1 0.57. **Table 3** presents the results for individual roles.

**Table 3.** Precision, recall and F1 for individual roles.

| Role | Precision | Recall | F1 |
|---|---|---|---|
| Analysis | .91 | .53 | .67 |
| Conceptualization | .75 | .50 | .60 |
| Experimenting | .22 | .80 | .34 |
| Study design | .77 | .60 | .67 |
| Coordination | 1.0 | .35 | .52 |
| Data collection | .58 | .56 | .57 |
| Paper drafting | .87 | .54 | .66 |
| Paper writing | .61 | .41 | .49 |
| Paper review | .95 | .50 | .66 |
| Paper revision | .93 | .31 | .46 |
| Paper reading | .81 | .85 | .83 |
| Literature review | .91 | .83 | .87 |
| Interpretation | .90 | .51 | .65 |

### 4.3 Error Analysis

We manually analyzed mistakes made by the extractor in the test set, and found two types: false positives that lower precision (a subject-role pair incorrectly present in the extracted output), and false negatives that that lower the recall (a correct subject-role pair missing from the extracted output). We identified three sources of errors (**Fig. 6**):

- Errors related to mention extraction from the text. That is, an incorrect mention is extracted, or a certain role mention is missing. These errors are responsible for 26% of false positives and 73% of false negatives.
- Errors appearing during role discovery, related to incorrect cluster merging. These errors result in the lack of roles *paper approving*, *supervision* and *funding acquisition* in the extractor's output and are responsible for 21% of false negatives. Classification errors, resulting in assigning an incorrect role to the tuple. These errors are responsible for 74% of false positives and 6% of false negatives.

The quality of the mention extraction has the biggest impact on the overall results, in particular recall. In a typical scenario, some mentions are missing from OpenIE output, which makes it impossible to extract specific subject-role pairs.

Incorrect tuples also affect the second cause of errors. For example, we observed that in many cases, Stanford's OpenIE tool extracts only one tuple from typical sentences similar to "All authors read and approve the final manuscript": ("all authors", "read", "the final manuscript"). In this case, the missing mention related to approving the manuscript resulted in the failure to discover this role in the corpus.

Finally, we observed that in some cases the classifier made the decision based on a single term such as "make", which does not carry enough information for a correct classification decision. Additional feature selection procedures for the classifier might result in better classification performance.

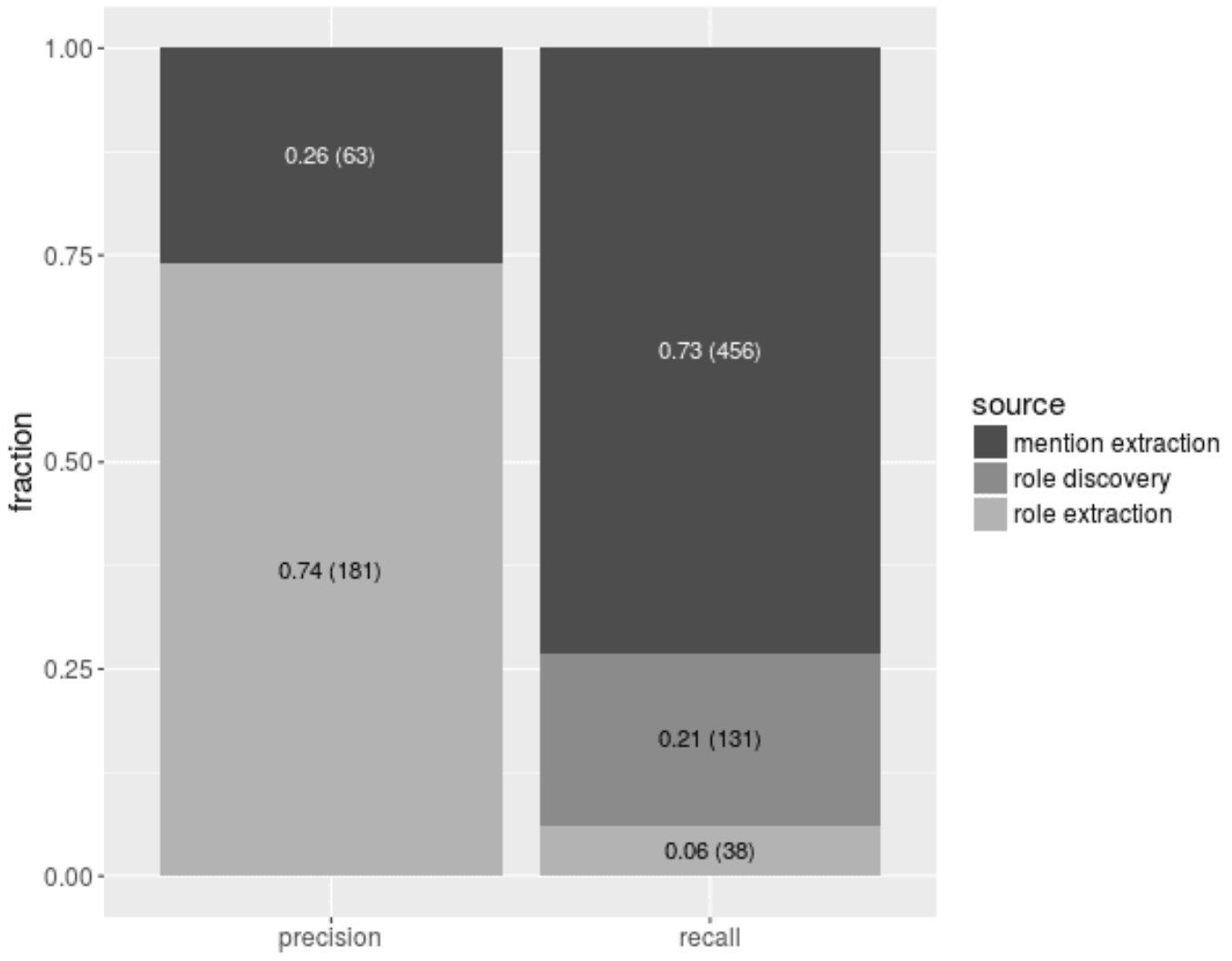

**Fig. 6.** The fraction of three error causes in types of errors (precision and recall errors).

## 5 Summary and Future Plans

In this paper, we presented a study of author contributions sections obtained from publications in biomedical disciplines. The results of our study include: 1) a set of roles discovered in the data in an unsupervised manner, and 2) a first prototype of a tool able to automatically extract the roles from the contributions section.

We semi-automatically discovered the following roles: *experimenting*, *analysis*, *study design*, *interpretation*, *conceptualization*, *paper reading*, *paper writing*, *paper review*, *paper drafting*, *coordination*, *data collection*, *paper revision* and *literature review*. Three discovered roles (*paper reading*, *literature review* and *interpretation*) are not included in the existing contributor roles taxonomy CRediT. The proposed automated role extractor is able to extract roles directly from the text with micro-averaged precision 0.68, recall 0.48 and F1 0.57.

Our plans for future work include: testing alternative mention extraction approaches and tools; testing alternative classification algorithms; and examining the relationships between author orderings, H-index and the nature of contributions in a larger corpus than used in previous analyses [9, 10].

## Acknowledgements

This research was conducted with the financial support of Enterprise Ireland and the European Regional Development Fund (ERDF) under Ireland's European Structural and Investment Funds Programme 2014-2020 under Grant Agreement No. CF/2017/0808-I at the ADAPT SFI Research Centre at Trinity College Dublin. The ADAPT SFI Centre for Digital Media Technology is funded by Science Foundation Ireland through the SFI Research Centres Programme and is co-funded under the European Regional Development Fund (ERDF) through Grant # 13/RC/2106.